\documentclass[lettersize,journal]{IEEEtran}
\usepackage{amsmath,amsfonts}
\usepackage{algorithmic}
\usepackage{array}
\usepackage[caption=false,font=normalsize,labelfont=sf,textfont=sf]{subfig}
\usepackage{textcomp}
\usepackage{stfloats}
\usepackage{url}
\usepackage{makecell}
\usepackage{multirow}
\usepackage{booktabs}
\usepackage{verbatim}
\usepackage{graphicx}

\usepackage{amssymb}
\usepackage{multicol}
\usepackage{pifont}
\usepackage{xcolor}
\usepackage{xpatch}
 \usepackage[pagebackref,breaklinks,colorlinks]{hyperref}

\usepackage[capitalize]{cleveref}
\crefname{section}{Sec.}{Secs.}
\Crefname{section}{Section}{Sections}
\Crefname{table}{Table}{Tables}
\crefname{table}{Tab.}{Tabs.}

\usepackage{color}
\usepackage{cite}
\def\BibTeX{{\rm B\kern-.05em{\sc i\kern-.025em b}\kern-.08em
    T\kern-.1667em\lower.7ex\hbox{E}\kern-.125emX}}
\usepackage{balance}

\begin{document}

\title{VicKAM: Visual Conceptual Knowledge Guided Action Map for Weakly Supervised Group Activity Recognition}

\author{Zhuming Wang\textsuperscript{1}, Yihao Zheng\textsuperscript{1}, Jiarui Li\textsuperscript{1}, Yaofei Wu\textsuperscript{1}, Yan Huang\textsuperscript{2}, Zun Li\textsuperscript{1}, Lifang Wu\textsuperscript{1}, Liang Wang\textsuperscript{2}

1 School of Information Science and Technology, Beijing University of Technology, Beijing, China\\
2 Institute of automation, Chinese Academy of Science, China\\
}
\maketitle

\begin{abstract}
   Existing weakly supervised group activity recognition methods rely on object detectors or attention mechanisms to capture key areas automatically. 
However, they overlook the semantic information associated with captured areas, which may adversely affect the recognition performance. 
In this paper, we propose a novel framework named Visual Conceptual Knowledge Guided Action Map (VicKAM) which effectively captures the locations of individual actions and integrates them with action semantics for weakly supervised group activity recognition.
It generates individual action prototypes from training set as visual conceptual knowledge to bridge action semantics and visual representations. Guided by this knowledge, VicKAM produces action maps that indicate the likelihood of each action occurring at various locations, based on image correlation theorem. 
It further augments individual action maps using group activity related statistical information, representing individual action distribution under different group activities, to establish connections between action maps and specific group activities. 
The augmented action map is incorporated with action semantic representations for group activity recognition.
Extensive experiments on two public benchmarks, the Volleyball and the NBA datasets, demonstrate the effectiveness of our proposed method, even in cases of limited training data. 
The code will be released later.
\end{abstract}

\section{Introduction}
\label{sec:intro}

Group Activity Recognition (GAR) is a critical task in computer vision focused on understanding collective behaviors exhibited by individuals within a group. It has garnered significant attention due to its wide-ranging applications, such as sports video analysis, surveillance, and social behavior understanding.
To effectively recognize group activities, it is essential to understand the actions of multiple individuals within a scene. 
Supervised methods~\cite{deng2016structure,ibrahim2016hierarchical,ibrahim2018hierarchical,li2017sbgar,qi2018stagnet,yan2018participation} explicitly align the visual features of individuals with action concepts through individual annotations, including bounding boxes during both training and testing phases, as well as action labels during training. 
Some approaches ~\cite{liu2021multimodal,tang2018mining,tang2019learning} incorporate action semantics to enhance the consistency between visual and semantic-level representations, and further improve recognition performance.
Although they demonstrate promising performance, annotating a large number of individual labels in real-world applications remains time-consuming and costly.

In recent years, weakly supervised group activity recognition approaches have gained more attention, particularly in scenarios where individual annotations are not available, at least in the testing set.
The intuitive idea ~\cite{bagautdinov2017social,zhang2019fast,yan2020sam}  is to first estimate individual locations through object detection and then perform group activity recognition via individual relation inference.
Yet these methods focus primarily on detecting individual locations, which cannot well establish a connection between the detected individuals and action semantics.
Some approaches~\cite{kim2022DF,chappa2023spartan,wu2024learning} employ attention mechanisms to automatically capture key areas related to group activities.
However, the areas they captured are not directly related to individuals or action semantics.
Therefore, although utilizing individual semantics has been demonstrated useful to enhance recognition performance, establishing a connection between visual representations and action semantics remains a challenge in weakly supervised group activity recognition.

\begin{figure}[!t]
  \centering
  \includegraphics[width = 0.97\linewidth]{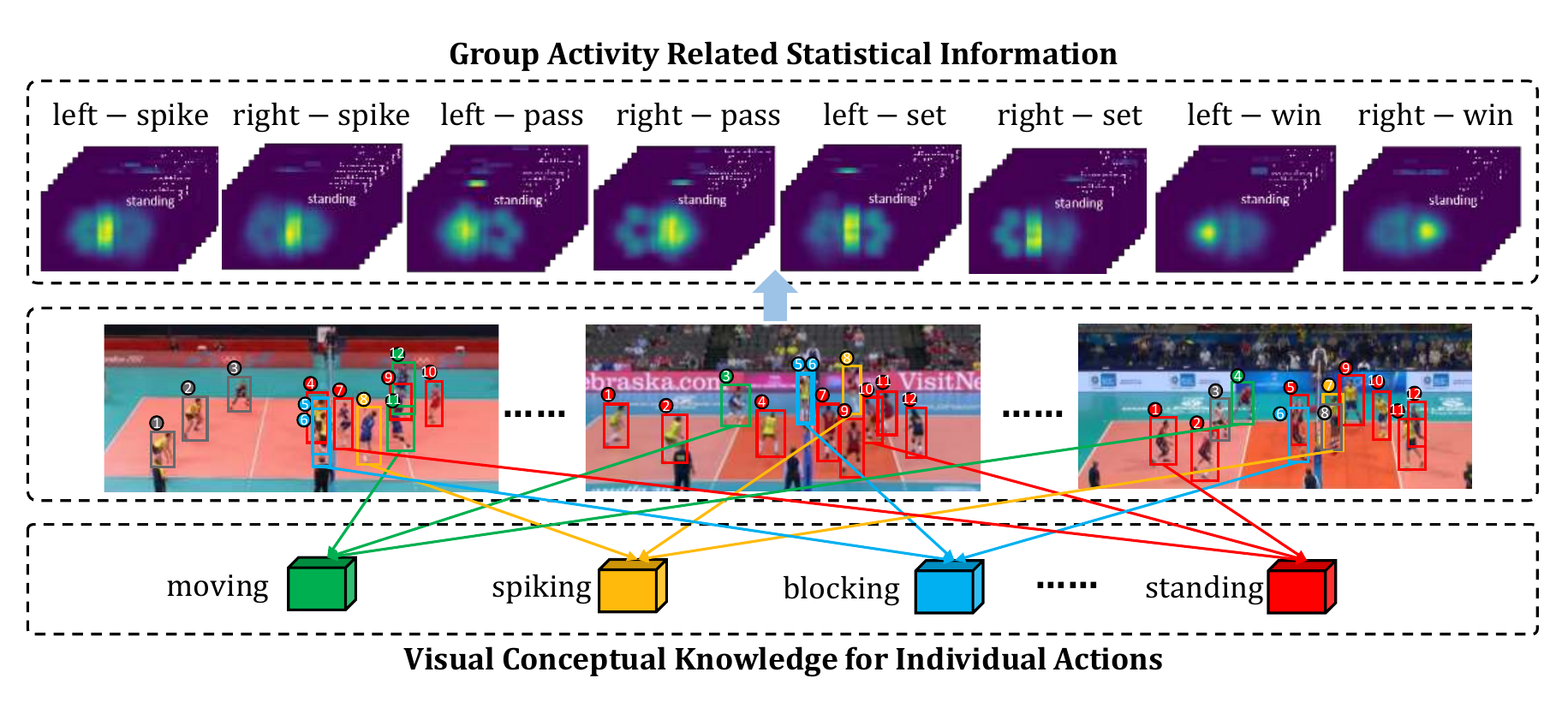}
  \caption{Demonstration of the visual conceptual knowledge and the group activity related statistical information. Individuals performing the same action exhibit similar visual features, so we average the visual features of many individuals engaged in that action to create its visual conceptual knowledge. Additionally, the distribution of individual actions usually follows a statistical pattern, which we summarize as statistical information corresponding to specific actions within different activities.}
  \label{fig_intro}
\end{figure}

In fact, we can draw inspiration from how the human brain correlates visual and semantic concepts: it does not require a large number of annotations during learning, but instead stores conceptual knowledge about objects and actions in a multi-modal aligned manner \cite{patterson2007you,bi2021dual,huang2022mack}. 
Inspired by this, we propose the visual conceptual knowledge for individual actions to connect their semantics with the visual representations of individuals performing them. 
As illustrated in \cref{fig_intro}, individuals performing the same action share similar visual features, facilitating the extraction of a general representation for each action from a set of samples. We define such general representations of individual actions as their prototypes, which reflect their visual conceptual knowledge.
Furthermore, a group activity is a collective expression of multiple individual actions, typically characterized by a specific pattern of action distribution \cite{wang2024knowledge}. These patterns can be summarized as activity-action relation maps providing statistical information about the distribution of individual actions in specific group activities.

In this paper, we propose a novel framework named Visual Conceptual Knowledge Guided Action Map (VicKAM) for weakly supervised group activity recognition, which effectively captures the locations of individual actions and integrates them with action semantics.
Specifically, VicKAM generates action maps that indicate the likelihood of each action occurring at various locations through image correlation theorem, guided by visual conceptual knowledge of individual actions.
It further establishes connections between action maps and specific group activities with activity related statistics and incorporates action semantic representations to improve group activity recognition.
We evaluate our approach on two public benchmarks, the Volleyball and the NBA datasets. Experimental results validate that our VicKAM performs favorably compared with the state-of-the-art methods.
The main contributions are summarized as follows:

\begin{itemize}
\item We propose an idea of creating visual conceptual knowledge to bridge the action semantics and visual representations. We further propose to generate individual action maps from visual conceptual knowledge to indicate the likelihood of each action occurring at various locations.

\item We propose a novel framework named Visual Conceptual Knowledge Guided Action Map for weakly supervised group activity recognition. It utilizes action maps to capture key areas connected with actions, and introduces activity related statistical information to augment the relation between actions relevant to specific activities. 

\item Experimental results from two public datasets demonstrate that our method achieves promising performance, especially when training data is limited.
\end{itemize}

\section{Related Work}
\label{sec:related}
\subsection{Fully Supervised Group Activity Recognition}
Existing GAR algorithms primarily rely on extracting visual information about individuals in the scene to infer group activities. 
Early works utilized hand-crafted features to recognize various activities through probabilistic graphical models~\cite{lan2012social,Amer2013,amer2014hirf,Li2016Multiview} or AND-OR models~\cite{amer2012cost,choi2013understanding,shu2015joint}.

With the rapid advancement of deep learning technology, GAR algorithms based on Convolutional Neural Networks have emerged as the primary focus of research.
Some approaches \cite{deng2016structure,ibrahim2018hierarchical,ibrahim2016hierarchical,li2017sbgar,qi2018stagnet,shu2017cern,wang2017recurrent,yan2018participation,ibrahim2016hierarchical,shu2017cern,CCG,li2019nonlocal} achieved satisfactory results in exploring the individual spatial-temporal relations within the scene based on Recurrent Neural Networks or Long Short-Term Memory structures. 
Recent developments in graph neural networks and transformers have improved the capability to model relations between individuals. 
Wu et al. ~\cite{ARG} devised an Actor Relation Graph (ARG) that constructed actor relation graphs to capture both appearance and position relations among actors.
Gavrilyuk et al. ~\cite{gavrilyuk2020actor} proposed an Actor-Transformer using RGB, optical flow, and pose features as input to model actors. 
Yuan et al. ~\cite{yuan2021learning} enhanced individual representations by incorporating global contextual information and aggregated the relation between individuals through a Spatial-Temporal Bi-linear Pooling module. 
Liu et al. ~\cite{liu2021multimodal} utilized individual action label embeddings to create a semantic graph that refines visual representations. 
Tang et al. ~\cite{tang2019learning} proposed to align individual visual representations with semantic representations derived from action labels through knowledge distillation. 

\begin{figure*}
    \centering
    \includegraphics[width=0.97\linewidth]{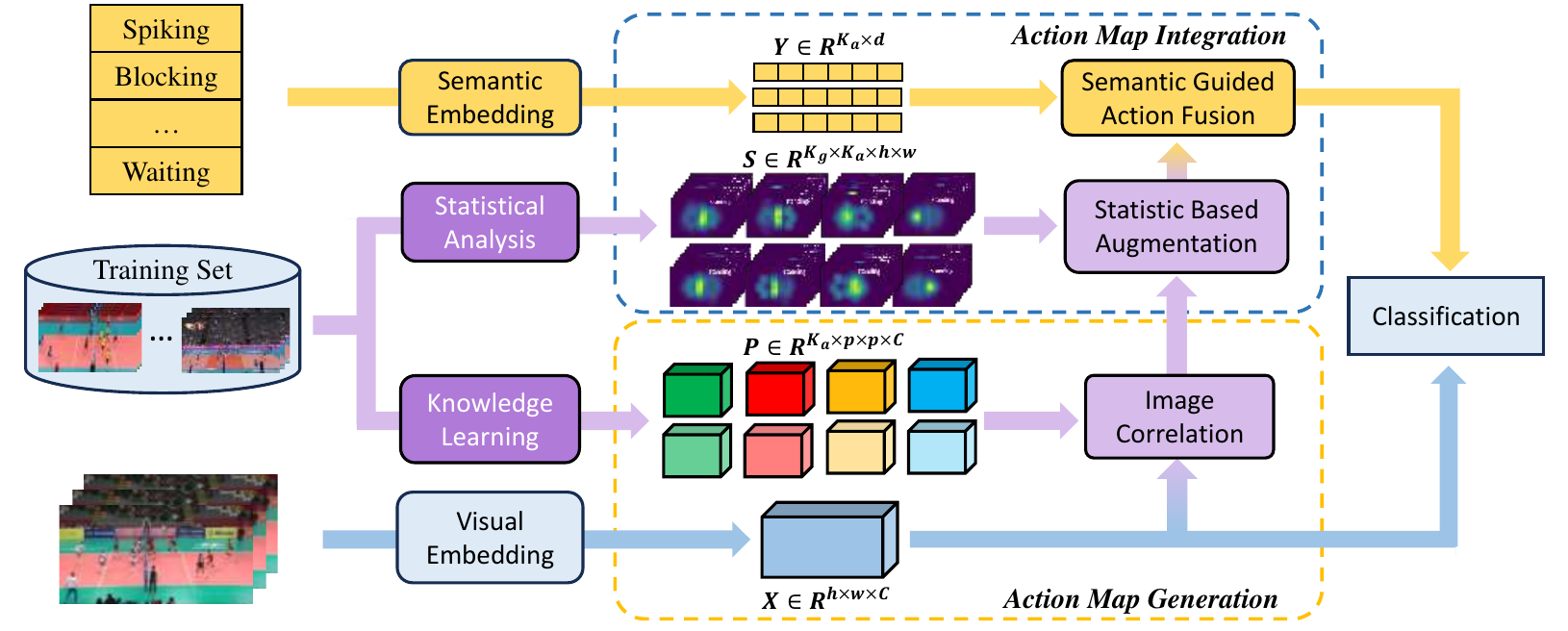}
    \caption{Overview of the proposed framework. 
    } 
    \label{fig_overview}
\end{figure*} 

\subsection{Weakly Supervised Group Activity Recognition}
Some algorithms aimed to overcome the limitations of individual annotations and explore group activity recognition in a weakly supervised setting.
Bagautdinov et al. ~\cite{bagautdinov2017social} simultaneously performed individual detection and feature extraction using a fully convolutional network, then fed the results into an RNN to recognize group activities in conjunction with individual actions.
Zhang et al. ~\cite{zhang2019fast} made the individual detection and weakly supervised group activity recognition collaborate in an end-to-end framework by sharing convolutional layers between them.
Yan et al. ~\cite{yan2020sam} addresses the issue of missing bounding boxes by generating actor boxes from detectors trained on external datasets and learning to prune irrelevant suggestions, thereby eliminating the need for actor-level labels during both training and inference. 

Apart from detector based algorithms, some methods utilize attention mechanisms to extract regions relevant to group activities.
Wu et al. \cite{wu2022active} utilized attention mechanisms to obtain masks that identify the spatial locations of scene activities and eliminate background information, using these masks as visual markers to construct spatial-temporal relations at different scales.
Kim et al. ~\cite{kim2022DF} proposed the Detector-Free method, which encodes the context of group activity as a set of visual embeddings, thereby bypassing the explicit target detection. 
Chappa et al. ~\cite{chappa2023spartan} employed self-distillation to learn frame-level and patch-level objectives in the latent space, aligning global spatio-temporal features from the entire sequence with local spatio-temporal features from the sampled sequence.
Wu et al. ~\cite{wu2024learning} embedded the specific label semantics to extract corresponding fine-grained information based on the hierarchy inherent in group-level labels, approaching GAR as a multi-label classification task.

However, these approaches lack an explicit connection between the visual information of individual actions and their semantic concepts, which has been demonstrated by fully supervised methods to be beneficial for recognizing group activities. 
To address these issues, we introduce visual conceptual knowledge that provides general visual representations of individual actions, and capture key areas based on actions through image correlation theorem.

\section{Method}
\label{sec:method}
\subsection{Overall Architecture}
Our framework is illustrated in \cref{fig_overview}. It first generates the individual action prototypes $\mathbf{P}$ and activity-action relation maps $\mathbf{S}$ from training data. 
Then, it performs action map generation for input videos guided by $\mathbf{P}$ using image correlation theorem. 
The obtained action maps are further refined through statistic based augmentation with $\mathbf{S}$, and integrate semantic representation $\mathbf{Y}$ for group activity recognition.

\subsection{Visual Conceptual Knowledge}
\begin{figure*}
    \centering
    \includegraphics[width=0.97\linewidth]{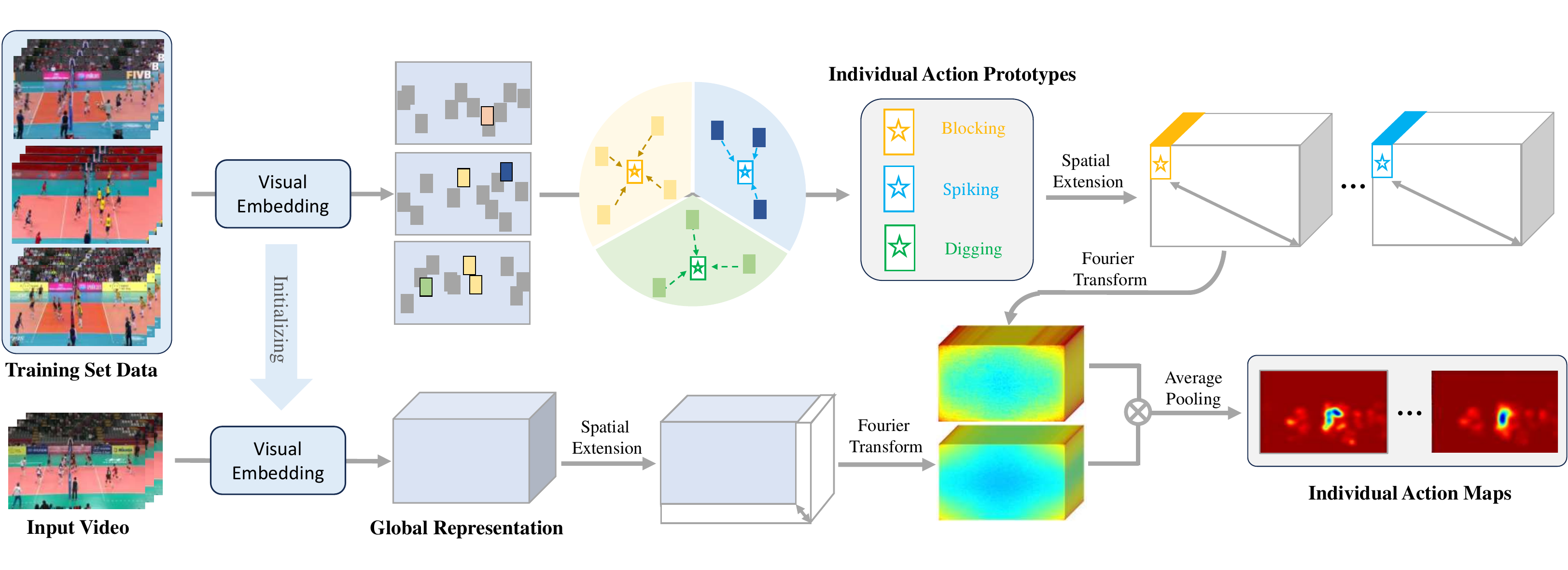}
    \caption{Illustration of Action Map Generation.
    } 
    \label{fig_action_mapping}
\end{figure*}

Prior to training the main framework, we leverage training set samples to explore visual conceptual knowledge of individual actions. As shown in ~\cref{fig_action_mapping}, we can gather a set of individuals performing a specific action, and create a general representation by averaging the visual features of all related individuals.
This step requires individual annotations, including both bounding boxes and action labels. 
 
Specifically, we employ a backbone followed by the RoiAlign~\cite{he2017mask} operation to extract the visual features of individuals.   
The output of the backbone is regarded as a global representation of the scene, sized $h \times w \times C$, which is further used to estimate the group activity through a simple classifier. Here, $h$ and $w$ denote the height and width of the feature resolution, respectively, and $C$ represents the number of feature channels.
The parameters of the backbone and the classifier are transferable to the main framework to initialize its video embedding blocks and global classifier as pre-trained weights. 
Meanwhile, the RoiAlign operation extracts visual features of individuals from the global representation, according to their bounding boxes. The visual features of individuals, sized $p \times p \times C$, are then processed by another classifier to identify individual actions.
Finally, we average the individual visual features according to their corresponding action categories, obtaining individual action prototypes $\mathbf{P} \in \mathbb{R}^{K_a \times p \times p \times C}$, where $K_a$ denotes the number of action categories.  

\subsection{Group Activity Related Statistical Information} 
The statistical information about the spatial distribution of individual actions can be easily summarized through statistical analysis on massive samples, if the scenarios involve regularized activity areas and individual positions. 
However, image samples from the same group activity may represent different camera views, leading to positional biases in the statistical spatial distribution.
Therefore, we perform camera view alignment ~\cite{wang2024knowledge}, which involves court line detection and affine projection, to align all samples to a uniform view with a resolution of $h \times w$. The aligned samples are used to generate activity-action relation maps.

We place the nearby regions of bottom center points of individual bounding boxes onto $K_a \times K_g$ sub-maps, based on the different actions each individual performs in various activities, where $K_g$ denotes the number of activity categories. We then concatenate these maps to create activity-action relation maps $\mathbf{S} \in \mathbb{R}^{K_g \times K_a \times h \times w}$, which indicate the likelihood of each individual action category occurring in specific spatial regions within a given activity.

\subsection{Action Map Generation}
Our proposed action maps are designed to indicate the likelihood of individuals performing various actions at different locations. They do not place excessive attention on a predefined number of individuals, but focus more on the action categories that individuals may perform and their occurring locations.
To achieve this, we incorporate image correlation theorem to facilitate the matching between the global representation of input videos and action prototypes.

Image correlation theorem defines an approach for analyzing image similarity by quantifying it through the weighted sum of pixel values in compared image pairs.
This necessitates point-wise calculations using a sliding window in the spatial domain, which can be time-consuming. The theory of Fourier transform offers an effective solution to this challenge.
As shown in ~\cref{fig_action_mapping}, given a video sequence, we leverage visual embedding blocks to extract its global representation $\mathbf{X} \in \mathbb{R}^{h \times w \times C}$.
Then, we apply zero-padding for spatial extension, adjusting the heights and widths of the resolution for global representation $\mathbf{X}$ and individual action prototypes $\mathbf{P}$ to $(h+p-1)$ and $(w+p-1)$. 
The generation of the action sub-map $\mathbf{M_k}$ corresponding to the $k$-th action category is formulated as:
\begin{equation}
\mathbf{M_k}=\mathbf{IFFT}\left(\mathbf{FFT}(\mathbf{X}) \times \mathbf{FFT}^*(\mathbf{P_k})\right)
\end{equation}
where $\mathbf{FFT}(\cdot)$ and $\mathbf{IFFT}(\cdot)$ denote the operations of Fast Fourier Transform and Inverse Fast Fourier Transform, while $\mathbf{FFT}^*(\cdot)$ represents the computation the complex conjugate of Fast Fourier Transform result. 
$M_k$ are further cropped to a resolution of $h \times w$ to match that of $\mathbf{X}$, and the $C$ channels are averaged to compress it to a single channel.
Each $M_k$ represents the occurrence patterns of the $k$-th action category in the input video. 
We then concatenate them as action maps $\mathbf{M}={\{M_k\}}^{K_a}_{k=1}$.

\subsection{Action Map Integration}
The action maps $\mathbf{M}$ provide a spatial description of the individual behaviors corresponding to the input video. We further utilized activity-action relation maps $\mathbf{S}$ as statistic based augmentation to refine $\mathbf{M}$, emphasizing the relation between actions relevant to specific activities.
In particular, we broadcast $\mathbf{M}$ to the size of $K_g \times K_a \times h \times w$ and then multiply it with $\mathbf{S}$. 
These refined action maps $\mathbf{\widehat{M}}$ incorporate spatially relevant action information provided by $\mathbf{S}$, along with co-occurrence patterns among actions associated with different activities.
Intuitively, they indicate whether the occurrence patterns of each action in the input video are consistent with the general patterns when a specific activity occurs.

Additionally, refined action maps reveal implicit associations among action semantics within various activities, determined by the sequence order of their internal sub-maps. To enhance consistency between sub-maps and action semantics, we introduce linguistic embeddings of action labels to create semantic representations that explicitly assign each sub-map.
We embed $K_a$ different action labels into a $d$ dimensional latent space to generate the semantic representations $\mathbf{Y} \in \mathbb{R}^{K_a \times d}$.

To integrate refined action maps, we first flatten the sub-maps $\mathbf{\widehat{M}_k}$ in $\mathbf{M}$ corresponding to the $k$-th group activity category into $K_a$ vectors, then encode these vectors into $D$ dimensional features.  
Next, we concatenate $\mathbf{Y}$ with these features and encode them into $D$ dimensional action features. Then, we stack and flatten them into a vector with a size of $K_a \times D$, and employ a Fully-connected layer to perform interaction among actions. 
The output $\mathbf{O_k}$ indicates the consistency between the input video and typical feature representation of the $k$-th activity.
Finally, we stack $\mathbf{O_k}$ associated with $K_g$ activities to obtain the group representation $\mathbf{O}={\{O_k\}}^{K_g}_{k=1}$.

\subsection{Training and Testing}
Our VicKAM is implemented in two stages: first, we utilize individual annotations to generate visual conceptual knowledge of individual actions and group activity related statistical information; then, we train the main framework without requiring any annotations, including bounding boxes or action labels.

In the first stage, we train the network using the following loss function:
\begin{equation}
    \mathcal{L}_{pre} = \mathcal{L}_{CE}(\hat{g},g) + \lambda_{pre} \sum\limits_{n=1}^N\mathcal{L}_{CE}(\hat{a}_{n},a_n)
\end{equation}
where $g$ and $a_n$ are the ground truth labels for group activities and individual actions, $\hat{g}$ and $\hat{a}_{n}$ are the predictions of activities and actions. $N$ is the number of individuals in the input video. $\mathcal{L}_{CE}$ represents the cross-entropy loss function. Parameter $\lambda_{pre}$ is a scalar used to balance the weight of two classification losses.

In the second stage, we train the main framework using the following loss function:
\begin{equation}
    \mathcal{L}_{main} = \mathcal{L}_{CE}(\hat{g}_s,g) + \lambda_{main} \mathcal{L}_{CE}(\hat{g}_o,g)
\end{equation}
where $\hat{g}_s$ denotes the prediction derived from the global representation $\mathbf{X}$, $\hat{g}_o$ denotes the prediction derived from the group representation $\mathbf{O}$.
$\lambda_{main}$ is a scalar used to balance the weight of two classification losses. We do not introduce any individual annotations at this stage.

In the test phase, we average $\hat{g}_s$ and $\hat{g}_o$ to produce the final prediction.
It is important to note that both visual conceptual knowledge of individual actions and group activity related statistical information can serve as general information transferred from the training data to the testing data. As a result, we directly use these components obtained from the training set without including any individual annotations in the testing set. 
Therefore, no individual bounding boxes or action labels are required during testing.

\section{Experiments}
\label{sec:experiments}

\begin{table}[h]
    \centering
    \begin{tabular}{@{}l@{\,}cc@{\,}c@{\,}cc@{}}
        \toprule
            \multirow{2}{*}{Scheme}  & \multirow{2}{*}{Backbone}  & \multicolumn{2}{c}{Train} & Test & \multirow{2}{*}{MCA$\uparrow$} \\
            \cmidrule{3-5}
            & & AL & BB & BB & \\
        \midrule
            \multicolumn{6}{c}{\bf{Fully Supervised}}\\
        \midrule
            HDTM~\cite{ibrahim2016hierarchical} & AlexNet & \ding{51} & \ding{51} & \ding{51} & 81.9\\
            HANs+HCNs~\cite{kong2018hierarchical} & GoogLeNet & \ding{51} & \ding{51} & \ding{51} & 85.1\\
            CERN~\cite{shu2017cern} & Vgg16   & \ding{51}  & \ding{51} & \ding{51} & 87.6\\
            stagNet~\cite{qi2018stagnet} & Vgg16 & \ding{51} & \ding{51} & \ding{51} & 89.3\\
            PRL~\cite{hu2020progressive} & Vgg16 & \ding{51} & \ding{51} & \ding{51} & 91.4\\
            AT~\cite{gavrilyuk2020actor} & I3D & \ding{51} & \ding{51} & \ding{51} & 91.4\\
            GINs~\cite{tang2020graph} & Vgg16 & \ding{51} & \ding{51} & \ding{51} & 91.7\\
            ARG~\cite{ARG} & Inception-v3 & \ding{51} & \ding{51} & \ding{51} & 92.5\\
            Ehsanpour et al~\cite{ehsanpour2020joint} & I3D & \ding{51} & \ding{51} & \ding{51} & 93.0\\
            STBiP~\cite{yuan2021learning} & Inception-v3 & \ding{51} & \ding{51} & \ding{51} & 93.3\\
            GroupFormer~\cite{li2021groupformer} & Inception-v3 & \ding{51} & \ding{51} & \ding{51} & 94.1\\
            Dual-AI~\cite{han2022dual} & Inception-v3 & \ding{51} & \ding{51} & \ding{51} & {94.4}\\
        \midrule
            \multicolumn{6}{c}{\bf{Weakly Supervised}}\\
        \midrule
            PoseConv3D~\cite{duan2022revisiting} & 3D-CNN &  & \ding{51} & \ding{51} & 91.3\\
            HIGCIN~\cite{yan2020higcin} & Resnet-18 &  & \ding{51}  & \ding{51} & 91.4\\
            Kong et al. ~\cite{kong2022spatio} & {Inception-v3} &  & \ding{51} & \ding{51} & 92.0\\
            DIN~\cite{yuan2021spatio} & Vgg16 &  & \ding{51} & \ding{51} & 93.6\\
            SSU~\cite{bagautdinov2017social} & {Inception-v3} & \ding{51} & {\ding{51}} &  & 87.1\\
            CRM~\cite{azar2019convolutional} & {I3D} & \ding{51} & {\ding{51}} &  & 92.1\\
            Zhang et al. ~\cite{zhang2019fast} & {ZFNet} &  & \ding{51} &  & 86.0\\
            ASPHRI~\cite{wu2022active} & {Inception-v3} &  & \ding{51} &  & 92.4\\
            PCTDM~\cite{yan2018participation} & ResNet-18 &  & &  & 80.5\\
            Detector-free~\cite{kim2022DF} & {ResNet-18} &  & &  & 90.5\\
            Wu et al. ~\cite{wu2024learning}  & ResNet-18  &  &  &  & 92.5\\
            SPARTAN~\cite{chappa2023spartan} & ViT & & & & 92.9\\
        \midrule
            Ours  & I3D  & \ding{51} & \ding{51} &  & 92.7\\
        \bottomrule
    \end{tabular}
    \caption{Comparison with the state-of-the-art on VD. `AL' denotes action labels, `BB' denotes bounding boxes.}
    \label{sota_vd}
\end{table}

\begin{table}[!t]
  \centering
  \begin{tabular}{lccc}
    \hline
    Method & Backbone & MCA$\uparrow$\\
    \hline
    ARG~\cite{ARG} & ResNet-18 & 59.0 \\
    AT~\cite{gavrilyuk2020actor} & ResNet-18 & 47.1 \\
    SACRF~\cite{pramono2020empowering} & ResNet-18 & 56.3\\
    DIN~\cite{yuan2021spatio} & ResNet-18 & 61.6 \\
    SAM~\cite{yan2020social} & ResNet-18 & 54.3 \\
    Dual-AI~\cite{han2022dual} & Inception-v3 & 51.5 \\
    Detector-free~\cite{kim2022DF} & ResNet-18 & 75.8 \\
    Wu et al. ~\cite{wu2024learning} & ResNet-18 & 75.8 \\
    \hline
    Ours & ResNet-18 & 69.8 \\
    \hline
  \end{tabular}
  \caption{Comparison with the state-of-the-art on NBA.}
    \label{sota_nba}
\end{table}

\subsection{Datasets and Metric} 
We conduct experiments on two widely used group activity recognition datasets,
namely the Volleyball Dataset (VD) and the NBA Dataset (NBA). 

{\bf{The Volleyball Dataset}} comprises 3,493 video clips for training and 1,337 for testing. It consists of 8 group activity categories, including left-spike, right-spike, left-set, right-set, left-pass, right-pass, left-win, and right-win. Each middle frame of the clips is annotated with individual bounding boxes and action categories, encompassing spiking, blocking, digging, setting, jumping, falling, moving, waiting, and standing. 

{\bf{The NBA Dataset}} contains 7624 training clips and 1548 testing clips. It consists of 9 group activity categories, including `2p-succ.', `2p-fail.-off.', `2p-fail.-def.',`2p-layup-succ.', `2p-layup-fail.-off.', `2p-layup-fail.-def.', `3p-succ.', `3p-fail.-off.', `3p-fail.-def.'. No individual annotations, including bounding boxes or action labels, are provided.

Following previous works, we use the widely recognized Multi-class Classification Accuracy (MCA) as the evaluation metric for experiments. 

\subsection{Implementation Details}
We select ten frames (the middle frame, 5 frames before it, and 4 frames after it) as input, for each video clip. 
The input resolutions are adjusted to 1280 $\times$ 720 to align with previous works~\cite{gavrilyuk2020actor,yuan2021learning,kim2022DF,han2022dual}.
The resolution $h \times w$ of global representation is 90$\times$160. The size of visual conceptual knowledge $p \times p \times C$ is 7$\times$7$\times$832. The $D$ and $d$ are set to 256 and 128, respectively.
We employed an inflated 3D ConvNet~\cite{i3d} as the backbone for VD and a ResNet-18 network for the NBA.
The classifier designed to predict $\hat{g}_s$ comprises an average pooling layer followed by a fully connected layer. In contrast, the classifier for predicting $\hat{g}_o$ is composed solely of a single fully connected layer.
The parameters $\lambda_{pre}$ and $\lambda_{main}$ are set to 1 and 3, respectively.
We utilize the Adam optimizer with an initial learning rate of 5e-4 for training both stages of our framework on VD. For the NBA, the learning rate is initially set to 5e-7, followed by a linear warm-up to 5e-5 over 5 epochs. After the 6th epoch, a linear decay of 1e-4 is applied. The batch size for both datasets is set to 4. 
Our framework is implemented using PyTorch and trained for 50 epochs on two NVIDIA GeForce RTX 3090 GPUs.

\subsection{Comparison with the State-of-the-Art}
\label{sota}
In this subsection, we present a comparative analysis of our VicKAM against SOTA methods. The comparison results are sourced from Wu et al. ~\cite{wu2024learning}.

\textbf{Comparisons on the Volleyball Dataset.}
The comparison results are presented in \cref{sota_vd}
Our framework achieves an MCA of 92.7\%, placing second among all comparisons. Although SPARTAN~\cite{chappa2023spartan} achieves a 0.2\% higher performance than ours, it relies on a Visual Transformer backbone and a self-distillation learning strategy, both of which require substantial computational resources.
Moreover, our scheme also outperforms several fully supervised schemes. 

\textbf{Comparisons on the NBA Dataset.}
The NBA Dataset does not provide individual annotations, therefore we introduce the MultiSports Dataset ~\cite{li2021multisports} to explore visual conceptual knowledge. 
This dataset provides the bounding box of the key actor in each video clip. 
We assign the key actors with `3-point', `2-point', and `rebound' action labels according to their corresponding video group activity label. 
These three categories of individual actions can roughly reflect the actions of key individuals involved in the `3p', `2p', and `layup' group activities within the NBA dataset. 
However, generating activity-action relation maps still remains challenging. Therefore, we did not employ statistic based augmentation in experiments on the NBA dataset.

The comparison results are presented in \cref{sota_nba}.
We analyze the reason for unsatisfactory performance on the NBA that, our proposed visual conceptual knowledge is designed to store action related information from the training environment and explicitly transfer it to the testing environment for guidance. However, the information obtained from the Multi-Sport dataset still exhibits significant differences from the data distribution in the NBA. As a result, the optimization direction of the global representation during the early stages of training may diverge considerably from the guidance provided by the visual conceptual knowledge, affecting its effectiveness. 

\begin{table}[htb]
    \centering
    \begin{tabular}{lcccc}
        \toprule
            \multirow{2}{*}{Scheme} & \multicolumn{4}{c}{Data Ratio} \\
        \cmidrule{2-5}
            & 10\% & 25\% & 50\% & 100\% \\
        \midrule
            \multicolumn{5}{c}{\bf{Fully Supervised}}\\
        \midrule
            PCTDM~\cite{yan2018participation}& 67.4 & 81.5 & 88.5 & 90.3 \\
            AT~\cite{gavrilyuk2020actor} & 67.7 & 84.2 & 88.0 & 90.0 \\
            HiGCIN~\cite{yan2020higcin} & 55.5 & 71.2 & 79.7 & 91.4\\
            ERN~\cite{ehsanpour2020joint} & 52.5 & 73.1 & 75.4 & 90.7\\
            ARG~\cite{ARG} & 80.2 & 87.9 & 90.1 & 92.3\\
            DIN~\cite{yuan2021spatio} & 71.7 & 84.1 &  89.9 & 93.1\\
            Dual-AI~\cite{han2022dual} & 85.5 & 89.7  & 92.7 & 94.4\\
        \midrule
            \multicolumn{5}{c}{\bf{Weakly Supervised}}\\
        \midrule
            Detector-free~\cite{kim2022DF} & 67.9 & 78.0 & 82.6 & 90.5 \\
            Wu et al. ~\cite{wu2024learning} & 80.1 & 83.7 & 86.2 & 92.5 \\
        \midrule
            Ours & 84.1 & 88.3 & 91.0 & 92.7 \\
        \bottomrule
    \end{tabular}
    \caption{Comparison with the state-of-the-art on VD under limited training data.}
    \label{limited}
\end{table}

\textbf{Experiment under Limited Training Data}
We present the comparison results of training on the Volleyball Dataset using 10\%, 25\%, 50\%, and 100\% of the samples in \cref{limited}.
Our framework achieves the best weakly supervised results under the limited training datasets, surpassing the existing best results by 4.0\%, 4.6\%, 4.8\%, and 0.2\%, respectively. 

This is attributed to our proposed visual conceptual knowledge and activity related statistical information. They store general visual and spatial distribution information about individual actions, which remains easily accessible and robust against interference, even with a low data ratio of 10\%.
These results clearly demonstrate the superiority of our method under limited training data.

\begin{table}[htb]
    \centering
    \begin{tabular}{lcc}
        \toprule
            Method & Params & FLOPs \\
        \midrule
            AT~\cite{gavrilyuk2020actor} & 29.6M & 305G \\
            SAM~\cite{yan2020social} & 25.5M & 304G \\
            SACRF~\cite{pramono2020empowering} & 53.7M & 339G \\
            ARG~\cite{ARG} & 49.5M & 307G \\
            DIN~\cite{yuan2021spatio} & 26.0M & 304G \\
            Detector-Free~\cite{kim2022DF} & 17.5M & 313G \\
            Wu et al. ~\cite{wu2024learning} & 17.8M & 309G \\
        \midrule
            Ours & 12.5M & 325G \\
        \bottomrule
    \end{tabular}
    \caption{Comparison of model complexity on NBA.}
    \label{complexity}
\end{table}

\textbf{Computational Complexity Analysis}
We compare the parameters and FLOPs of our scheme with the state-of-the-art on NBA dataset in ~\cref{complexity}. 
It can be observed that our framework achieves comparable performance while maintaining the lowest number of parameters (12.5M) and competitive FLOPs (325G).

\begin{table}[htb]
    \centering
    \begin{tabular}{cccccc}
        \toprule
            Model & Backbone & Act. & Stat. & Sem. & MCA$\uparrow$\\
        \midrule
            (A) & \ding{51} & & & & 91.1 \\
            (B) & \ding{51} & \ding{51} &  &  & 91.8 \\
            (C) & \ding{51} & \ding{51} & \ding{51} & & 92.2 \\
            (D) & \ding{51} & \ding{51} &  & \ding{51} & 92.3 \\
        \midrule
            Ours & \ding{51} & \ding{51} & \ding{51} & \ding{51} & 92.7 \\
        \bottomrule
    \end{tabular}
    \caption{Experiment results of ablation studies on VD. `Act.' refers to the use of action maps; `Stat.' refers to incorporation of activity-action relation maps for statistic based augmentation; `Sem.' refers to introducing semantic representations into action map integration.}
    \label{ablation_components}
\end{table}

\begin{table}[htb]
    \centering
    \begin{tabular}{cccccc}
            \toprule
                Resolution & 3$\times$3 & 5$\times$5 & 7$\times$7 & 9$\times$9 & 11$\times$11 \\
            \midrule
                MCA$\uparrow$ & 91.8 & 92.3 & 92.7 & 92.6 & 92.1 \\ 
            \bottomrule
        \end{tabular}
    \caption{Impact of individual action prototype resolutions.}
    \label{ablation_resolution_knowledge}
\end{table}

\begin{table}[htb]
    \centering
    \begin{tabular}{cccccc}
            \toprule
                Resolution & 7$\times$7 & 11$\times$11 & 15$\times$15 & 19$\times$19 & 23$\times$23 \\
            \midrule
                MCA$\uparrow$ & 91.9 & 92.3 & 92.6 & 92.7 & 92.3 \\
            \bottomrule
        \end{tabular}
    \caption{Impact of marked region resolutions in activity-action relation maps.}
    \label{ablation_region_spmaps}
\end{table}

\subsection{Ablation Studies}
\label{ablation}
We conduct ablation studies on the Volleyball Dataset to investigate the contribution of components in our scheme. The corresponding results are shown in \cref{ablation_components}.

\textbf{Effect of Action Maps.}
We compare two ablation models as follows: (A) directly leverage global representation $\mathbf{X}$ to produce the final prediction; (B) utilizes action maps $\mathbf{M}$, without introducing statistic based augmentation or semantic representations. 
Leveraging action maps improves the MCA from 91.1\% to 91.8\%, compared to directly using global representation for classification. Regarding the base model (A) predicts activities solely based on global visual information, action maps emphasize the importance of capturing key areas related to action concepts. 

\textbf{Effect of Statistic Based Augmentation.}
We design model (C) which adds statistic based augmentation to model (B). It further improves the MCA from 91.8\% to 92.2\%, demonstrating the benefit of incorporating distribution information of actions associated with activities.

\textbf{Effect of Semantic Representation.}
We compare model (D), which incorporates semantic representations into the action map integration operation, with model (B). This integration improves the MCA from 91.8\% to 92.3\%, validating the effectiveness of explicitly enhancing the consistency between individual action maps and their semantical concepts.

\textbf{Impact of Individual Action Prototype Resolutions.}
The resolution $p \times p$ of individual action prototypes is pre-defined. We conduct ablation studies to investigate the impact of different resolutions on VD. 
As shown in ~\cref{ablation_resolution_knowledge}, the framework achieves similar performance when the resolution is set to 5$\times$5, 7$\times$7 and 9$\times$9.
We set the resolution to 7$\times$7 in VicKAM to achieve optimal performance.

\textbf{Impact of Marked Region Resolutions.}
We conduct ablation studies to investigate the impact of different resolutions of marked regions when generating activity-action relation maps on VD.
As shown in ~\cref{ablation_region_spmaps}, it is evident that setting the region resolution to 15$\times$15 and 19$\times$19 yielded similar performance. However, further reducing or increasing the resolution resulted in performance degradations.

This could be attributed to the fact that when the region resolution is too large, the activity-action relation maps become overly dense, diminishing the distinctions between points and causing the entire network to resemble the ablation model (D). Conversely, when the region resolution is too small, the activity-action relation maps become excessively sparse, making the action maps overly sensitive to positional distribution.
We choose 19$\times$19 as the marked region resolution for VicKAM.

\begin{figure}[!t]
  \centering
  \includegraphics[width = 0.97\linewidth]{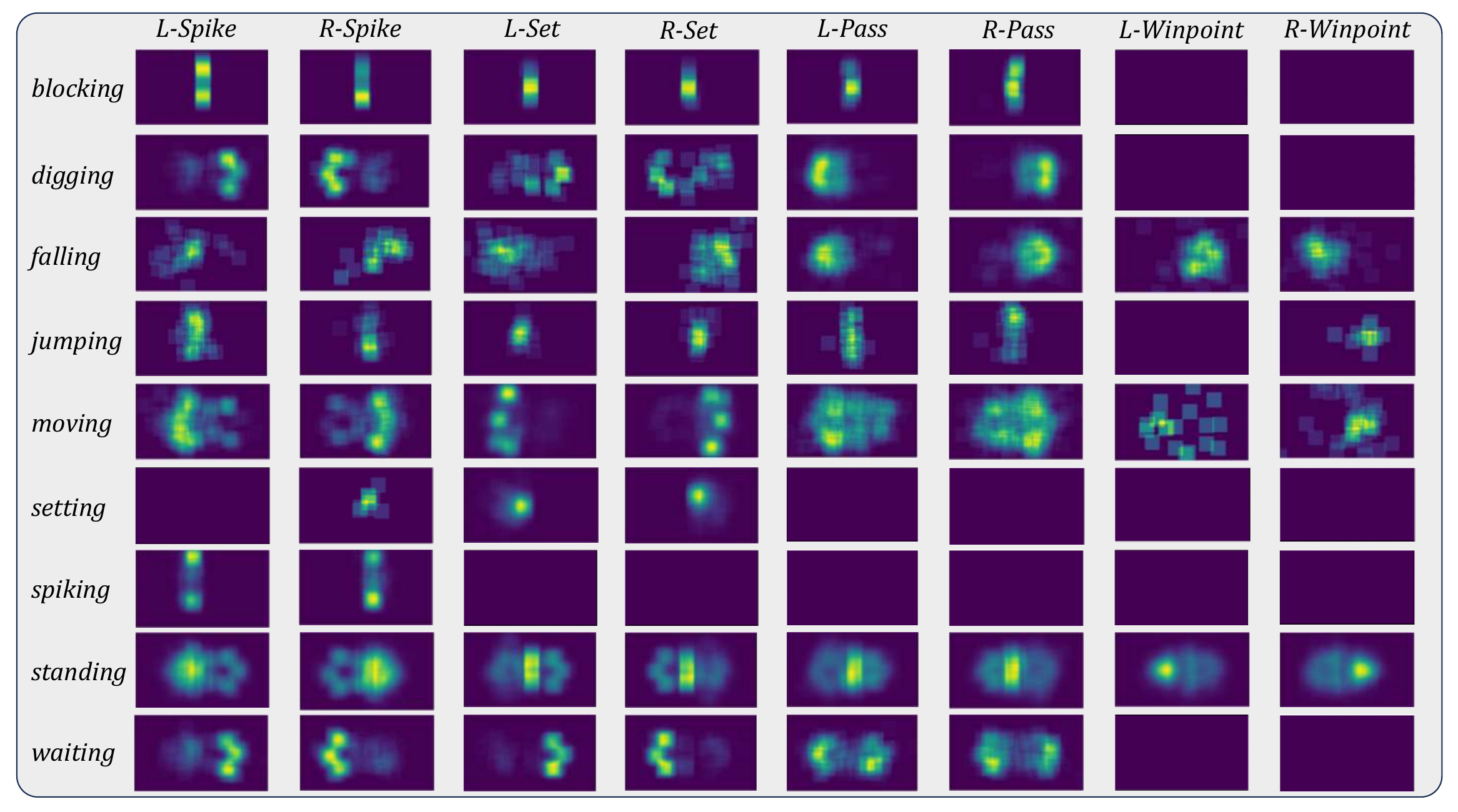}
  \caption{Visualizations of Activity-Action Relation Maps.}
  \label{fig_sp}
\end{figure}

\begin{figure}[!t]
  \centering
  \includegraphics[width = 0.97\linewidth]{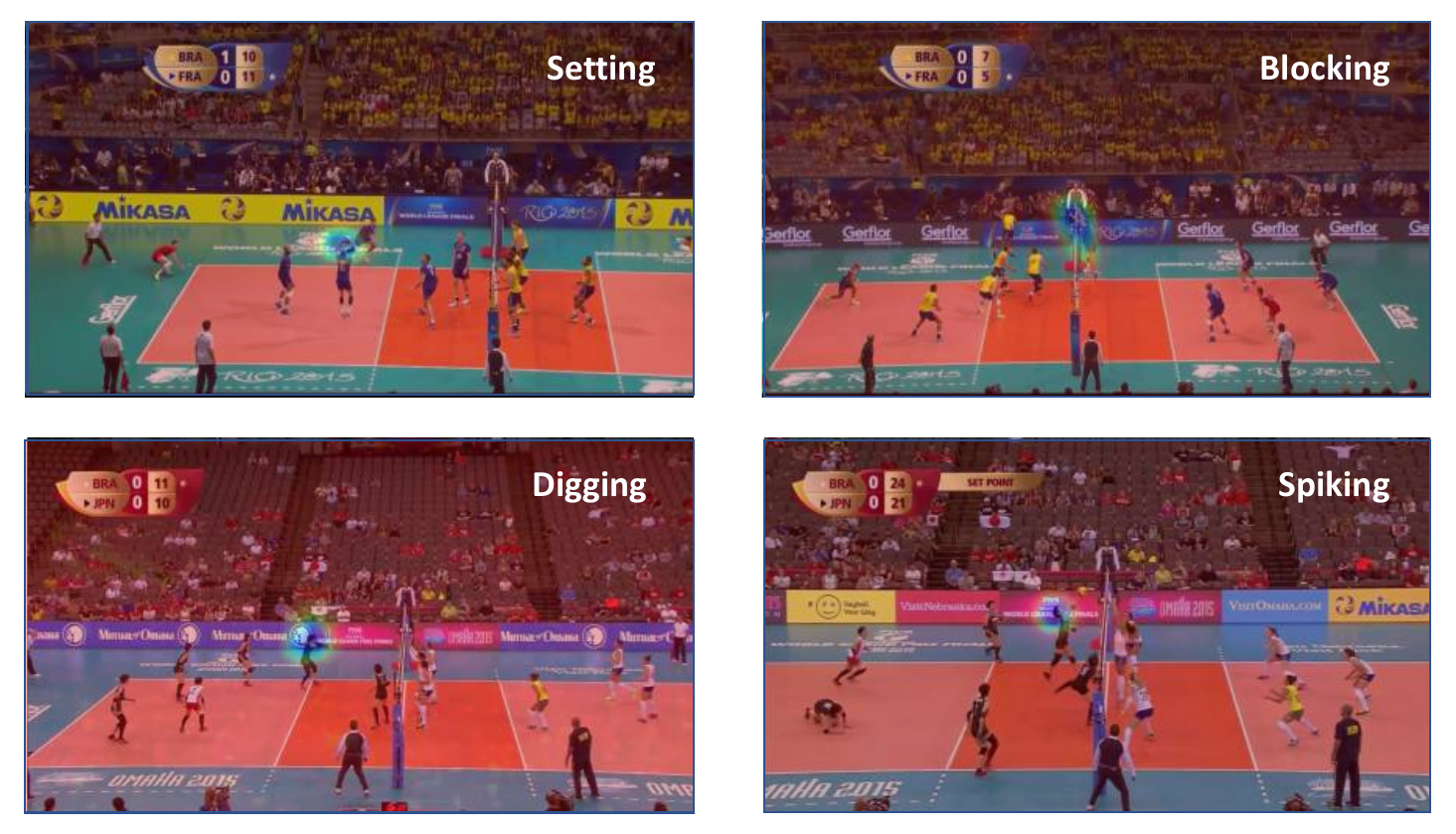}
  \caption{Visualizations of Action Maps.}
  \label{fig_actionmap}
\end{figure}

\begin{figure}[!t]
  \centering
  \includegraphics[width = 0.97\linewidth]{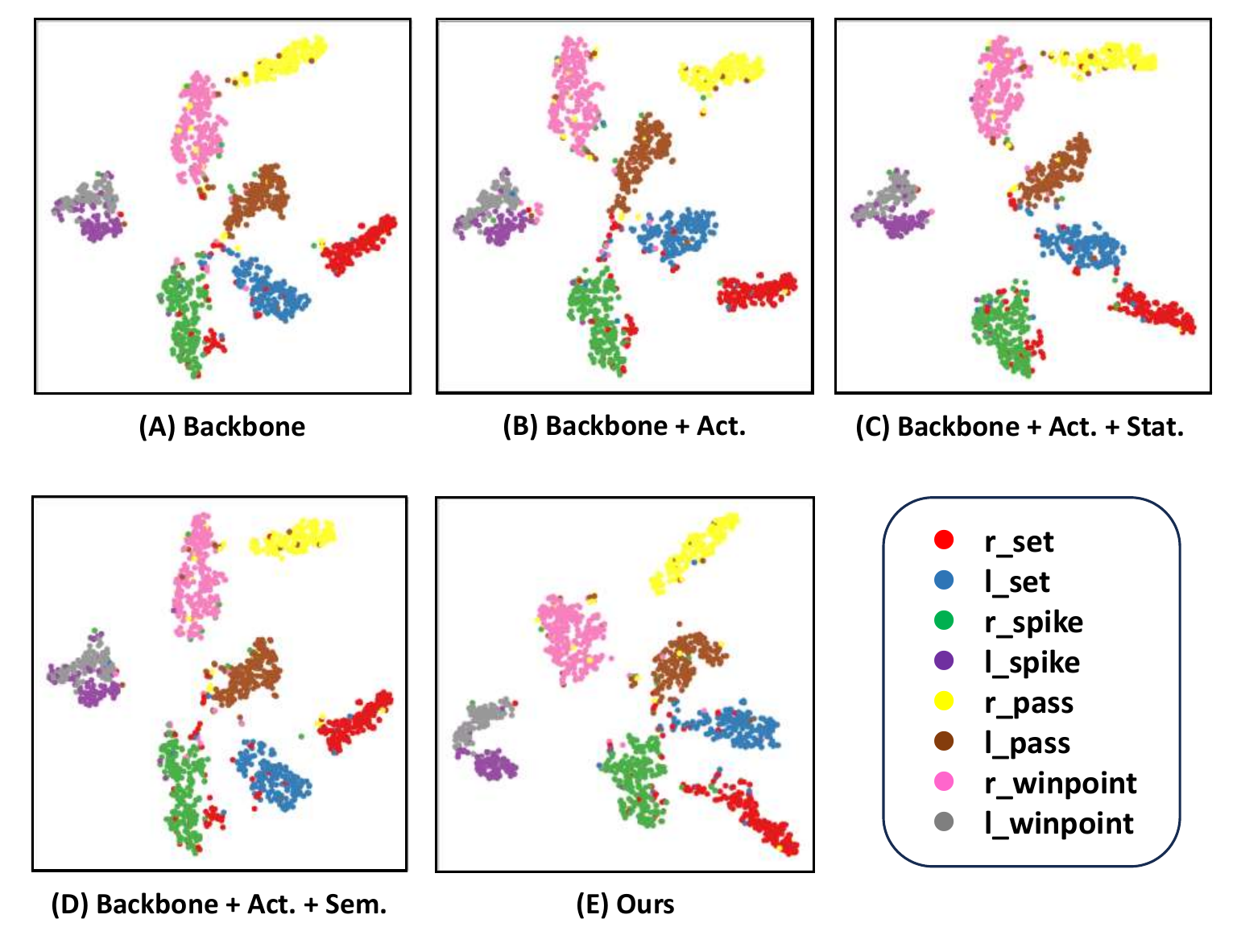}
  \caption{T-SNE visualization of the learned representations.}
  \label{fig_tsne}
\end{figure}

\subsection{Visualization}
\textbf{Activity-Action Relation Maps.}
We present visualizations of activity-action relation maps in \cref{fig_sp}. We can observe the symmetry of different actions within the left and right subcategories of the same activity. 

We can also observe the distribution of the setting action under the right-spike activity, which is inconsistent with reality. We believe this discrepancy arises from mislabeling in the dataset samples.

Additionally, the distribution areas for spiking, blocking and jumping are generally shifted upward. This occurs because individuals performing these actions typically are airborne, resulting in differences between the visual and actual spatial positions.

\textbf{Action Maps.}
In \cref{fig_actionmap}, we illustrate the visualization of action maps obtained from image correlation on VD. It is evident that the action maps effectively highlight areas where the corresponding actions took place.

\textbf{The t-SNE visualization}
We visualize the distribution of learned representations by t-SNE~\cite{van2008visualizing} on the VD.
From ~\cref{fig_tsne} (A), it is evident that the features corresponding to different activities do not effectively differentiate from one another when only the backbone is utilized. In particular, the left-spike (purple) and left-winpoint (gray) categories exhibit significant overlap.
Introducing action maps, activity related statistical information, or semantic representations, as shown in ~\cref{fig_tsne} (B-D), can help disperse the clusters of different group categories, but they still do not resolve the overlap issue. 
However, as shown in ~\cref{fig_tsne} (E), our complete framework exhibits a superior ability to differentiate all activity features, including left-spike and left-winpoint.

\section{Conclusion}
\label{sec:conclusion}
In this work, we propose a novel idea of generating visual conceptual knowledge to bridge action semantics and visual representations of input videos. 
Building on this, we propose a framework named Visual Conceptual Knowledge Guided Action Map for weakly supervised group activity recognition. It utilizes visual conceptual knowledge to identify key areas associated with actions, derives action maps for input videos through image correlation theorem, and incorporates activity related statistical information to emphasize the relation between actions relevant to specific activities. 
Comprehensive experiments conducted on two public datasets demonstrate the promising performance of our scheme, particularly in scenarios with limited training data.

We believe our work provides valuable insights for future research in this field. However, there are still limitations in our approach to address, such as the necessity of individual annotations for exploring visual conceptual knowledge and its suboptimal performance in guiding action map generation across different domains.

\bibliographystyle{IEEEtran}
\bibliography{main}

\end{document}